\documentclass[10pt,journal,compsoc]{IEEEtran}
\ifCLASSOPTIONcompsoc
  \usepackage[nocompress]{cite}
\else
  \usepackage{cite}
\fi
\ifCLASSINFOpdf
\else
\fi

\usepackage{hyperref}       % hyperlinks
\usepackage{url}            % simple URL typesetting
\usepackage{booktabs}       % professional-quality tables
\usepackage{amsfonts}       % blackboard math symbols
\usepackage{nicefrac}       % compact symbols for 1/2, etc.
\usepackage{microtype}      % microtypography
\usepackage{amsmath}
\usepackage{amssymb}
\usepackage{subcaption}
\usepackage{graphicx}

\begin{document}
\title{$K$-Shot Contrastive Learning of Visual Features with Multiple Instance Augmentations}

\author{Haohang~Xu,~%\IEEEmembership{Member,~IEEE,}
        Hongkai~Xiong,~%\IEEEmembership{Fellow,~OSA,}
        and~Guo-Jun~Qi,~\IEEEmembership{Senior Member,~IEEE}% <-this % stops a space
\IEEEcompsocitemizethanks{
\IEEEcompsocthanksitem H. Xu and H. Xiong are with Shanghai Jiaotong University.
\IEEEcompsocthanksitem G.-J. Qi was with the Seattle Cloud Lab, Futurewei Technologies, Bellevue, WA 98004.\protect\\
E-mail: guojunq@gmail.com
}% <-this % stops an unwanted space
}


\markboth{Preprint}%
{Shell \MakeLowercase{\textit{et al.}}: Bare Demo of IEEEtran.cls for Computer Society Journals}



\IEEEtitleabstractindextext{%
\begin{abstract}
In this paper, we propose the $K$-Shot Contrastive Learning (KSCL) of visual features by applying multiple augmentations to investigate the sample variations within individual instances.  It aims to combine the advantages of {\em inter-instance discrimination} by learning discriminative features to distinguish between different instances, as well as {\em intra-instance variations} by matching queries against the variants of augmented samples over instances.
%through a combination of the significant factors of variations in the augmented samples.
Particularly, for each instance, it constructs an instance subspace to model the configuration of how the significant factors of variations in $K$-shot augmentations can be combined to form the variants of augmentations. Given a query, the most relevant variant of instances is then retrieved by projecting the query onto their subspaces to predict the positive instance class. This generalizes the existing contrastive learning that can be viewed as a special one-shot case. An eigenvalue decomposition is performed to configure instance subspaces, and the embedding network can be trained end-to-end through the differentiable subspace configuration. Experiment results demonstrate the proposed $K$-shot contrastive learning achieves superior performances to the state-of-the-art unsupervised methods.
\end{abstract}

}

\maketitle

\IEEEdisplaynontitleabstractindextext

% \fi
\IEEEpeerreviewmaketitle

\IEEEraisesectionheading{\section{Introduction}}\label{sec:intro}


\IEEEPARstart{U}{nsupervised}  learning of visual features has attracted wide attentions as it provides an alternative way to efficiently train very deep networks without labeled data that are often expensive to collect \cite{hua2008online,qi2008correlative}. Recent breakthroughs in this direction focus on two categories of methods: contrastive learning \cite{oord2018representation,chen2020simple,he2019momentum} and transformation prediction \cite{gidaris2018unsupervised,zhang2019aet,qi2019avt,noroozi2016unsupervised}, among many alternative unsupervised methods such as generative adversarial networks \cite{goodfellow2014generative,qi2020loss,qi2018global,zhao2018adversarial}, and auto-encoders \cite{masci2011stacked,hinton2011transforming}.

The former category \cite{oord2018representation,chen2020simple,he2019momentum,bachman2019learning,henaff2019data,hjelm2018learning,tian2019contrastive} trains a network based on a self-training task by distinguishing between different {\em instance classes} each containing the samples augmented from the same instance. Such a contrastive learning problem seeks to explore the {\em inter-instance discrimination} to perform unsupervised learning.  On the contrary, the other category of transformation prediction methods \cite{gidaris2018unsupervised,zhang2019aet,qi2019avt,noroozi2016unsupervised} train a deep network by predicting the transformations used to augment input instances.  It attempts to explore the {\em intra-instance variations} under multiple augmentations to learn the feature embedding.

A good visual representation ought to combine both advantages of the inter-instance discrimination and the intra-instance variations \cite{qi2020small,oord2018representation,zhang2019aet,wang2012recommending}. In particular, the feature embedding should not only capture the significant intra-instance variations among augmented samples from each instance, as well as discern the distinction between instances by considering their potential variations to enable the inter-instance discrimination. In other words, the inter-instance discrimination should be performed by matching a query against all potential variants of an instance.
%intra-instance variations should capture the distribution of augmented samples from each instance, and the inter-instance discrimination forces the learned feature to discern the distinction between instances.
%under multiple augmentations
%While the inter-instance discrimination forces the learned feature to discern the distinction between instances, .
To this end, we propose a novel $K$-shot contrastive learning as a first attempt to combine their strengths, and we will show that most of existing contrastive learning methods are a special one-shot case.

In particular, we apply multiple augmentations to transform each instance, resulting in an {\em instance subspace} spanned by the augmented samples.  Each instance subspace learns significant factors of variations from the augmented samples, which configures how these factors can be linearly combined to form the variants of the instance. Then given a query, the most relevant sample of variant for each instance is retrieved by projecting the query onto the associated subspace \cite{hua2008online,shu2016image}. After that, the inter-instance discrimination is conducted by assigning the query to the instance class with the shortest projection distance.
%is then projected onto individual instance subspaces, and the inter-instance discrimination is performed by retrieving the similar samples belonging to the instance class with the longest projection.
An eigenvalue decomposition is performed to configure each instance subspace with the orthonormal eigenvectors as its basis. This configuration of instance subspaces is non-parametric and differentiable, allowing an end-to-end training of the embedding network \cite{chang2014factorized} through back-propagation.

Experiment results demonstrate that the proposed $K$-Shot Contrastive Learning (KSCL) can consistently improve the state-of-the-art performance on unsupervised learning. Particularly, with the ResNet50 backbone, it improves the top-1 accuracy of the SimCLR and the MoCo v2 to 68.8\% on ImageNet over 200 epochs.  It also reaches a higher top-1 accuracy of $71.4\%$ over 800 epochs than the baseline SimCLR and the rerun MoCo v2. For the sake of fair comparison, all these improvements are achieved with the same experiment settings such as network architecture, data augmentation, training strategy and the version of deep learning framework and libraries. The consistently improved performances with the same model settings suggest the proposed KSCL can serve as a generic plugin to further increase the accuracy of contrastive learning methods on downstream tasks.

%This demonstrates the proposed algorithm consistently outperforms through more shots of instance augmentations.

The remainder of the paper is organized as follows. We will review the related works in Section~\ref{sec:related},  and present the proposed $K$-shot contrastive learning in Section~\ref{sec:appr}. Implementation details will be depicted in Section~\ref{sec:imp}. We will demonstrate the experiment results in Section~\ref{sec:exp}, and conclude the paper in Section~\ref{sec:concl}.

\section{Related Works}\label{sec:related}
In this section, we review the related works to the proposed K-Short Contrastive Learning (KSCL) in the following four areas. A more comprehensive review of related methods on unsupervised models can be found in \cite{qi2020small}.

\subsection{Contrastive Learning}
Contrastive learning \cite{oord2018representation} was first proposed to learn unsupervised representations by maximizing the mutual information between the learned representation and a particular context. It usually focused on the context of the same instance to learn features by discriminating between one example from the other in an embedding space \cite{wu2018unsupervised,he2019momentum,chen2020simple}. For example, the instance discrimination has been used as a pretext task by distinguishing augmented samples from each other in a minimatch \cite{chen2020simple}, over a memory bank \cite{wu2018unsupervised}, or a dynamic dictionary with a queue \cite{he2019momentum}.  The comparison between the augmented samples of individual instances was usually performed on a pairwise basis.  The state-of-the-art performances on contrastive learning have relied on a composite of carefully designed augmentations \cite{chen2020simple} to prevent the unsupervised training from utilizing side information to accomplish the pretext task.  This has been shown necessary to reach competitive results on downstream tasks.

\subsection{Transformation Prediction}

Transformation prediction \cite{zhang2019aet,gidaris2018unsupervised} also constitutes a category of unsupervised methods in learning visual embeddings. In contrast to contrastive learning that focuses on inter-instance discrimination, it aims to learn the representations that equivary against various transformations \cite{qi2019avt,qi2019learning,gao2020graphter}.  These transformations are used to augment images and the learned representation is trained to capture the visual structures from which these transformations can be recognized.  It focuses on modeling the intra-instance variations from which variants of an instance can be leveraged on downstream tasks such as classification \cite{zhang2019aet,qi2019learning,gidaris2018unsupervised}, object detection \cite{gidaris2018unsupervised,qi2019learning}, semantic segmentations  on images \cite{gidaris2018unsupervised,qi2016hierarchically} and 3D cloud points \cite{gao2020graphter}. This category of methods provide an orthogonal perspective to contrastive learning based on inter-instance discrimination.

\subsection{Few-Shot Learning}

From an alternative perspective, contrastive learning based inter-instance discrimination can be viewed as a special case of few-shot learning \cite{peng2019few,jamal2019task,xu2019flat,qin2018prior,qi2016joint}, where each instance is a class and it has several examples augmented from the instance.  The difference lies that the examples for each class can be much abundant since one can apply many augmentations to generate an arbitrary number of examples. Of course, these examples are not statistically independent as they share the same instance. Based on this point of view, the non-parametric instance discrimination \cite{wu2018unsupervised} and thus several perspective works \cite{he2019momentum,chen2020simple} can be viewed as an extension of the weight imprinting \cite{qi2018low} by initialing the weights of each instance class with the embedded feature vector of an augmented sample, resulting in the inner product and cosine similarity used in these algorithms \cite{wu2018unsupervised,he2019momentum,chen2020simple}.  Such a surprising connection between the non-parametric instance discrimination and the few-shot learning may open a new way to train the contrastive prediction model. In this sense, the proposed $K$-shot contrastive learning generalizes the few-shot learning by imprinting the orthonormal basis of an instance subspace with the embeddigns of augmented samples from the instance.

\subsection{Capsule Nets}

The length of a vectorized feature representation has been used in capsule nets pioneered by Hinton et al.~\cite{sabour2017dynamic,zhang2018cappronet}. In capsule nets, a group of neurons form a capsule (vector) of which the direction represents different instantiation that equivaries against transformations and the length accounts for the confidence that a particular class of object is detected. From this point of view, the projected vector of a query example to an instance subspace in this paper also carries an analogy to a capsule.  Its direction represents the instantiated configuration of how $K$-shot augmentations from the instance are linearly combined to form the query, while its length gives rise to the likelihood of the query belonging to this instance class, since a longer projection means a shorter distance to the subspace.  This idea of using projections onto several capsule subspaces each corresponding to a class has shown promising results by effectively training deep networks \cite{zhang2018cappronet}.

\section{The Approach}\label{sec:appr}

In this section, we define a $K$-shot contrastive learning as the pretext task for training unsupervised feature embedding with Multiple Instance Augmentations (MIAs).

\subsection{Preliminaries on Contrastive Learning}
Suppose we are given a set of $N$ unlabeled instances $\mathcal X \triangleq\{\mathbf x_n\}$ in a minibatch (e.g., in the SimCLR \cite{chen2020simple}) or from a dictionary (e.g., the memory bank in non-parametric instance discrimination \cite{wu2018unsupervised} and the dynamic queue in the MoCo \cite{he2019momentum}). Then the contrastive learning can be formulated as classifying a query example $\mathbf x$ into one of $N$ instance classes each corresponding to an instance $\mathbf x_n$.

The goal is to learn a deep network embedding each instance $\mathbf x_n$ and the query $\mathbf x$ to a feature vector $\mathbf v_n$ and $\mathbf v$. Then the probability of the embedded query $\mathbf v$ belonging to an instance class $n$ is defined as
\begin{equation}\label{eq:closs}
p(n|\mathbf v) = \dfrac{\exp({\rm sim}(\mathbf v_n,\mathbf v)/\tau)}{\sum_{i=1}^N\exp({\rm sim}(\mathbf v_i,\mathbf v)/\tau)}
\end{equation}
where a similarity measure ${\rm sim}(\cdot,\cdot)$ (e.g., cosine similarity) is defined between two embeddings, and $\tau$ is a positive temperature hypermeter. When the query $\mathbf v$ is the embedding of an augmented sample from $\mathbf x_n$, $p(n|\mathbf v)$ gives rise to the probability of a relevant embedding $\mathbf v_n$ being successfully retrieved from the instance class $n$. One can minimize the contrastive loss called InfoNCE in \cite{oord2018representation} resulting from the negative log-likelihood of the above probability over a dictionary to train the embedding network.

%A negative log-likelihood of $p(n|\mathbf v)$ is minimized over a minimatch or a memory bank to train the embedding network, with $\mathbf v$ drawn from the instance class $n$ (i.e., it is the embedded feature of an augmented sample from $\mathbf x_n$).

The idea underlying the contrastive learning approach is a good representation ought to help retrieve the relevant samples from a set of instances $\mathcal X$ given a query $\mathbf x$. For example, the SimCLR \cite{chen2020simple} has achieved the state-of-the-art performance by applying two separate augmentations to each instance in a minibatch. Then, given a query example, it views the sample augmented from the same instance as the positive example, while treating those augmented from the other instances as negative ones.
Alternatively, the MoCo \cite{he2019momentum} seeks to retrieve relevant samples from a dynamic queue separate from the current minibatch.
%The dynamic query is kept being updated from the previous minibatches, and a momemtum its key encoder is updated  in a
%given a query through a momentum update of embeddings in the queue.
Both are based on the similarity between a query and a candidate sample to train the embedding network, which can be viewed as one-shot contrastive learning as explained later.

%Neither SimCLR nor MoCo has applied multiple augmentations in a minibatch to reveal the underlying intra-instance variations for each instance class.
However, the discrimination between different instance classes not only relies on their inter-instance similarities, but also is characterized by the distribution of augmented samples from the same instance, i.e., the intra-instance variations.
While existing contrastive learning methods explore the {\em inter-instance discrimination} to predict instance classes, we believe the {\em intra-instance variations} also play an indispensable role. Thus we propose $K$-shot contrastive learning by matching a query against the variants of each instance in the associated instance subspace spanned by $K$-shot augmentations.

\subsection{$K$-Shot Multiple Instance Augmentations}

\begin{figure*}[t]
%\captionsetup[subfigure]{singlelinecheck=false}
    \centering
    \begin{subfigure}[c]{0.99\textwidth}
        \includegraphics[width=\textwidth]{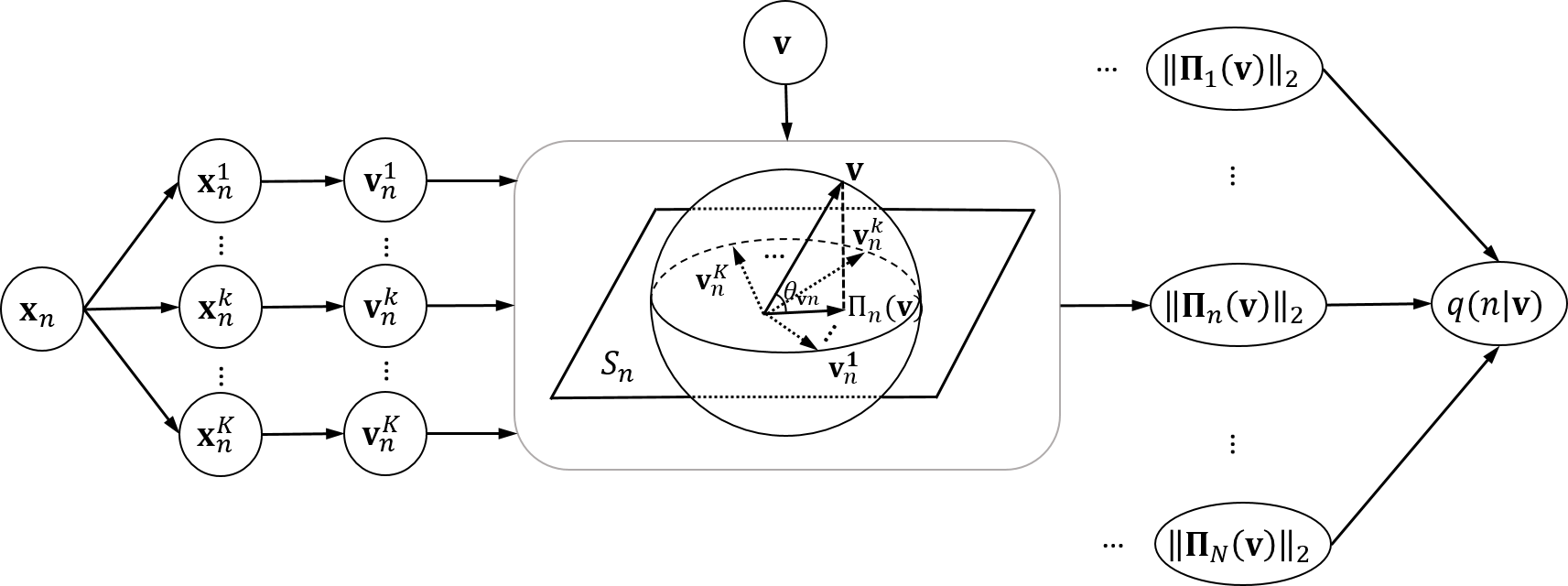}
        %\caption{Auto-Encoding Data (AED)}
    \end{subfigure}
    \caption{The pipline of the proposed $K$-Shot Contrastive Learning (KSCL). For each instance $\mathbf x_n$, an instance subspace $\mathcal S_n$ is spanned by the $\ell_2$-normalized embeddings $\{\mathbf v_n^k\}$ of $K$-shot augmentations $\{\mathbf x_n^k\}$ on a unit hyper-sphere. A given query embedding $\mathbf v$ of unit length is projected onto the subspace of each instance, resulting in the projection length $\|\Pi_n(\mathbf v)\|_2$ to measure the probability $q(n|\mathbf v)$ of the query belonging to the associated instance class. The projection length also gives the cosine similarity of the acute angle $\theta_{\mathbf vn}$ between the query vector $\mathbf v$ and the instance subspace $\mathcal S_n$.
    %In implementing the KSCL, an orthogonal basis for each instance subspace is derived with an eigenvalue decomposition, which is a differentiable function of the embedded augmentations.  This enables an end-to-end training of the embeddings by back-propagating the cross entropy errors from $q(n|\mathbf v)$'s.
    }\label{fig:kscl}
\end{figure*}

Let us consider a $K$-Shot Contrastive Learning (KSCL) problem. Suppose that $K$ different augmentations are drawn and applied to each instance $\mathbf x_n$, resulting in $K$ augmented samples $\mathbf x_n^k$ and their embeddings $\mathbf v_n^k$ for $k=1,\cdots,K$.

As aforementioned, the information contained in $K$-shot augmentations provides important clues to distinguish between different instance classes. Comparing a query against each augmented sample individually fails to leverage such intra-instance variations, since the most relevant sample of variant could be a combination of rather than individual factors of variations.  Therefore, we are motivated to explore the intra-instance variations through a linear subspace spanned by the augmented samples of each instances. Given a query, the most relevant instance is retrieved by projecting it onto the closest subspace.

%A naive application of contrastive learning may draw positive pairs from $K$ augmented samples, and apply the MoCo and SimCLR to train the embedding network.
%
%However, there are several disadvantages with such an approach. First, it neglects the distribution of augmented samples per instance class without fully exploring the intra-instance variances among augmented samples. The information contained in $K$-shot augmentations indeed provides important clues to distinguish between different instance classes, which cannot be well captured by only comparing a query against individual augmented samples. Second, randomly sampling positive pairs to learn feature embeddings is much inefficient as only pair-wise similarities are involved in calculating $p(n|\mathbf v)$.

%Specifically, in a $K$-shot contrastive learning,
As illustrated in Figure~\ref{fig:kscl}, consider the embeddings $\{\mathbf v_n^k\}$ of $K$-shot augmentations for an instance $\mathbf x_n$.  These embeddings are normalized to have a unit length and thus reside on the surface of a unit hyper-sphere. Meanwhile, they span an {\em instance subspace} $\mathcal S_n$ in the ambient feature space.
%We translate  the origin of $\mathcal S_n$ to the mean $\mathbf m_n={\rm mean}(\{\mathbf v_n^k|k=1,\cdots,K\})$ of the $K$ embeddings. This enables the subspace $\mathcal S_n$ to focus on the local variations of augmented samples relative to the mean vector, making it more discriminative to fine-grained variations for individual instances.
Then, the projection of the query $\mathbf v$ (of a unit length) onto the instance subspace is $\Pi_n (\mathbf v)$, and the projection distance of the query from $\mathcal S_n$ becomes
\begin{equation}\label{eq:dist}
D_n(\mathbf v) = \|\mathbf v-\Pi_n (\mathbf v)\|_2=\sqrt{\|\mathbf v\|_2^2 - \|\Pi_n (\mathbf v)\|_2^2}
\end{equation}
where $[\mathbf v-\Pi_n (\mathbf v)]$ is normal to $\mathcal S_n$, and the second equality follows from $[\mathbf v-\Pi_n (\mathbf v)]~\bot~ \Pi_n (\mathbf v)$, since the normal vector should be orthogonal to all vectors within the subspace.
As the embedding $\mathbf v$ has a constant unit length $\|\mathbf v\|_2=1$, minimizing the projection distance is equivalent to maximizing its projection length $\|\Pi_n (\mathbf v)\|_2$.

Let $\theta_{\mathbf vn}$ be the acute angle between the query $\mathbf v$ and the instance subspace $\mathcal S_n$. Then we have $\|\Pi_n (\mathbf v)\|_2=\cos(\theta_{\mathbf vn})$, i.e., the projection length can be viewed as the cosine similarity between the query and the {\em whole} instance subspace. Compared with the cosine similarity between individual embeddings of instances used in literature \cite{chen2020simple,he2019momentum,chen2020improved}, it aims to learn a better representation by discriminating different instance subspaces containing the variations of sample augmentations.

% a better representation can be learned to discriminate different instances by exploring the intra-instance variations in the $K$-shot contrastive learning.

%This differs from the cosine similarity between two individual vectors used in many existing contrastive learning methods neglecting the variations of samples spanning the underlying instance subspace.
%In the next section, we will show how to efficiently compute the projection length $\|\Pi_n (\mathbf v)\|_2$.

%By projecting the query $\mathbf v$ onto this translated subspace, we obtain a projected vector $\Pi_n (\mathbf v)$, and its distance

Now we can define the probability of $\mathbf v$ belonging to an instance class $n$
\begin{equation}\label{eq:kcloss}
p(n|\mathbf v) = \dfrac{\exp(\|\Pi_n (\mathbf v)\|_2/\tau)}{\sum_{m=1}^N\exp(\|\Pi_m (\mathbf v)\|_2/\tau)}
\end{equation}
%where $\|\cdot\|_2$ measures the projection length of the query in each subspace.
Then the KSCL seeks to train the embedding network by maximizing the loglikehood of the above probability over mini-batches to match a query against the correct instance.  Particularly, given a query $\mathbf v$ of a unit norm, its projection length achieves its maximum if $\mathbf v$ belongs to $\mathcal S_n$, i.e., it is a linear combination of $K$-shot augmentations $\{\mathbf v_n^k\}$.
%In this case, it also minimizes the projection distance as shown in Eq.~(\ref{eq:dist}).
In other words, it matches the query against all linear combinations of the augmented samples from each instance $\mathbf x_n$, and retrieves the most similar one by projecting the query onto the instance subspace with the shortest distance.

%In the later, we will only involve the significant factors of variations per instance to improve the robustness in retrieving relevant samples.

%We will minimize the negative likelihood of the above probability in a minibatch \cite{} or over a memory bank \cite{} to train the embedding network.  In this paper, we will adopt the memory bank like in MoCo to learn the feature representation.
%A set of positive pairs of samples are formed between queries and the relevant samples

\section{Implementations}\label{sec:imp}
In this section, we discuss the details to implement the proposed $K$-Shot Contrastive Learning (KSCL) model.

\subsection{Projection onto Instance Subspace via Eigenvalue Decomposition}
Mathematically, there is a close-form solution to the projection $\Pi_n(\mathbf v-\mathbf m_n)$ onto the instance subspace $\mathcal S_n$ spanned by $K$-shot augmentations $(\mathbf v_n^k)$'s.
%To avoid the notation clutter, we will consider the translated version of $\mathbf v$ and $\mathbf v_n^k$ in the following
%$$
%\mathbf v \leftarrow \mathbf v - \mathbf m_n, ~ \mathbf v_n^k\leftarrow\mathbf v_n^k-\mathbf m_n
%$$
Suppose there exists an othonormal basis for $\mathcal S_n$ denoted by the columns of a matrix $\mathbf W_n$, the projection $\Pi_n(\mathbf v)$ of a feature vector $\mathbf v$ can be written as $\mathbf W_n \mathbf W_n^\intercal \mathbf v$.

Since we have $\Pi_n(\mathbf v_n^k)=\mathbf v_n^k$ with $\{\mathbf v_n^k\}$ spanning $\mathcal S_n$, the problem of finding $\mathbf W_n$ can be formulated by minimizing the following projection residual
\begin{equation}\label{eq:eig}
\min_{\mathbf W_n^\intercal\mathbf W_n=\mathbf I}~\sum_{k=1}^K\|\mathbf v_n^k-\Pi_n(\mathbf v_n^k)\|={\rm tr}(-\mathbf W_n^\intercal\boldsymbol\Sigma_n\mathbf W_n + \boldsymbol\Sigma_n)
\end{equation}
where $\boldsymbol\Sigma_n=\mathbf V_n\mathbf V_n^\intercal$, with $\mathbf V_n\triangleq\left[\mathbf v_n^1, \cdots,\mathbf v_n^K\right]$ containing the embeddings of the $K$ augmented samples in its columns.

%By stacking them into the columns of a matrix $\mathbf V_n=\left[\mathbf v_n^1, \cdots,\mathbf v_n^K\right]$, we obtain a covariance matrix $\boldsymbol\Sigma_n=\mathbf V_n\mathbf V_n^\intercal$ per instance $n$, which is positive semi-definite.
After conducting an eigenvalue decomposition on the positive-definite matrix $\boldsymbol\Sigma_n$,  the eigenvectors corresponding to the largest $K$ eigenvalues give rise to an orthonormal basis $\mathbf W_n$ of the associated instance subspace, which minimizes (\ref{eq:eig}). %Because the number $K$ of augmented samples is usually smaller than the dimension of the embedded feature vector, the rank of $\boldsymbol\Sigma_n$ is no larger than $K$, and thus the residual of Eq.~(\ref{eq:eig}) will become zero.

Since the eigenvalue decomposition is differentiable, the embedding network can be trained end-to-end through the error back-propagation. However, like the other contrastive learning methods \cite{he2019momentum,chen2020improved}, the errors will only be back-propagated through the embedding network of queries to save the computing cost.

%where $\mathbf W_n$ is a $K\times K$ orthonormal matrix and its column vectors form the orthonormal basis of the subspace $\mathcal S_n$.
%By stacking this basis into the columns of a matrix $\mathbf W_n$, the projection $\Pi_n(\mathbf v)$ can be written as $\mathbf W_n \mathbf W_n^\intercal \mathbf v$ \footnote{The number $K$ of augmented samples per instance is usually smaller than the dimension of the embedded feature vector, and it is not hard to show that these $K$ eigenvectors $\mathbf W_n$ minimizes $\sum_{k=1}^K\|\mathbf v_n^k-\Pi_n(\mathbf v_n^k)\|={\rm tr}(-\mathbf W_n^\intercal\boldsymbol\Sigma_n\mathbf W_n + \boldsymbol\Sigma_n)$ as an orthornormal basis that spans the instance subspace.}. Since the eigenvalue decomposition is differentiable, the embedding network can be trained end-to-end through the error back-propagation.

\subsection{Most Significant Inter-Instance Variations}
Usually, we only consider a smaller number of eigenvectors, corresponding to the largest $L(<K)$ eigenvalues that account for the most significant factors of variations among $K$-shot augmentations.  This ignores the remaining $(K-L)$ minor factors of intra-instance variations that may be incurred by noisy augmentations. It also results in a thinner projection matrix $\mathbf {\widetilde W}_n$ than $\mathbf W_n$, and the projection length becomes $\|\Pi_n(\mathbf v)\|_2 = \|\mathbf {\widetilde W}_n\mathbf v\|_2$. Thus, we will only need to store and update $\mathbf {\widetilde W}_n$ in the KSCL.
%, instead of individual augmented samples as in other contrastive learning methods \cite{chen2020simple,he2019momentum,wu2018unsupervised}.

 %we may only consider the $L$ largest components for each subspace, while ignoring those corresponding to the smallest $(K-L)$ components of instance augmentations. Thus, only the first $L$ columns of $\mathbf W_n$ corresponding to the $L$ largest eigenvalues  will be preserved, resulting in $\mathbf {\widetilde W}_n$ to approximately span $\mathcal S_n$. Then, we have the length of the projection as $\|\Pi_n(\mathbf v)\|_2 = \|\mathbf {\widetilde W}_n\mathbf v\|_2$.

In practice, rather than setting $L$ to a prefixed number, we will choose $L$ such as the largest $L$ eigenvalues cover a preset percentage of total eigenvalues. The more percentage of total eigenvalues are preserved, the smaller the projection residual is in Eq.~(\ref{eq:kcloss}); when $L\geq K$, the residual vanishes. This allows a distinct number of eigenvectors per instance to flexibly model various degrees of variations among $K$-shot augmentations.

\subsection{One-Shot Contrastive Learning when $K=1$}
It is not hard to see that the cosine similarity used in SimCLR and MoCo is a special case when $K=1$, i.e., they are {\em one-shot contrastive learning} of visual embeddings. When $K=1$, there is a single augmented sample $\mathbf v_n$ per instance. Its instance subspace $\mathcal S_n$ collapses to a vector $\mathbf v_n$. Since $\mathbf v_n$ is $\ell_2$-normalized to have a unit length in the SimCLR and the MoCo, the projection length of a query $\mathbf v$ to this single vector becomes $|\mathbf v_n^\intercal \mathbf v|$. This is the cosine similarity  between two vectors used in existing contrastive learning methods \cite{chen2020simple,he2019momentum,wu2018unsupervised} up to an absolute value.

\section{Experiments}\label{sec:exp}
In this section, we perform experiments to compare the KSCL with the other state-of-the-art unsupervised learning methods.

\subsection{Training Details}
To ensure the fair comparison with the previous unsupervised methods \cite{chen2020simple,he2019momentum,chen2020improved}, in particular SimCLR \cite{chen2020simple} and MoCo v2 \cite{chen2020improved}, we follow the same evaluation protocol with the same hyperparameters.

Specifically, a ResNet-50 network is first pretrained on 1.28M ImageNet dataset \cite{deng2009imagenet} without labels, and the performance is evaluated by training a linear classifier upon the fixed features. We report top-1 accuracy on the ImageNet validation set with a single crop to $224\times 224$ images. The momentum update with the same size of dynamic queue, the MLP head, the data augmentation (e.g., color distortion and blur augmentation) are also adopted for the sake of fair comparison with the SimCLR and MoCo v2. We adopt the same temperature $\tau=0.2$ in \cite{chen2020improved} without exhaustively searching for an optimal one, yet still obtain better results. This demonstrates the proposed KSCL can be used as a universal plugin to consistently improve the contrastive learning with no need of further tuning of existing models.  We will evaluate the impact of $K$ and the percentage $\rho$ of preserved eigenvalues on the performance later.

\subsection{Results on ImageNet Dataset}

\begin{table*}
\centering
\caption{The top-1 accuracy of different models on ImageNet. The ResNet-50 backbone was unsupervisedly pretrained with two-layer MLP head by applying the same combination of enhanced data augmentations used in SimCLR including stronger color distortion and blurring for a fair comparison. The proposed KSCL is trained with $K=5$ and $\rho=40\%$. The top-1 accuracy is obtained by training a single-layer linear classifier upon the pretrained features. }\label{tb:imagenet}\vspace{2mm}
 \begin{tabular}{c|cc|c} \toprule
Model   &epochs&batch size&top-1 accuracy\\ \midrule
SimCLR\cite{chen2020simple} & 200 & 256 & 61.9\\
SimCLR (baseline)\cite{chen2020simple}  & 200 & 8192 & 66.6\\
MoCo v1\cite{he2019momentum} & 200&256& 60.5\\
MoCo v2 (rerun)\cite{chen2020improved} & 200 & 256 &67.5\\\midrule
Proposed KSCL & 200 & 256 & {\bf 68.8}\\\midrule\midrule
 \multicolumn{4}{c}{Results under more epochs of unsupervised pretraining}\\\midrule
SimCLR (baseline)\cite{chen2020simple} & 1000 & 4096 & 69.3\\
MoCo v2 (rerun) \cite{chen2020improved} & 800 & 256 & 70.6 \\\midrule
Proposed KSCL & 800 & 256 & {\bf 71.4}\\ \bottomrule
\end{tabular}
\end{table*}

Table~\ref{tb:imagenet} compares the top-1 accuracy of the proposed KSCL with that of SimCLR and MoCo on the ImageNet.
We make a direct comparison between the KSCL and the MoCo v2  by running both on the same hardware platform with the same set of software such as CUDA10, pytorch v1.3 and torchvision 1.1.0 (used in the data augmentation that plays a key role in the contrastive learning). With 200 epochs of pretraining, the same top-1 accuracy has been achieved on MoCo v2. However, its rerun result over 800 epochs is slightly lower than the reported result (71.1\%) in literature \cite{chen2020improved}, and this may be due to different versions of deep learning frameworks and drivers that could cause variations in the model performance.

Table~\ref{tb:imagenet} shows that, after unsupervised pretraining of the KSCL with $200$ epochs and a batch size of 256, the KSCL achieves a top-1 accuracy of 68.8\% with $K=5$ augmentations and $\rho=40\%$ of preserved eigenvalues. It is worth noting that a larger batch size is often required to sufficiently train the SimCLR while the other models such as KSCL and MoCo maintain a long dynamic queue as the dictionary. By viewing the SimCLR with a larger batch size of $8,192$ as a baseline, the KSCL makes a much larger improvement of $2.1\%$ than the MoCo v2 ($0.9\%$) on the SimCLR baseline under $200$ epochs. The KSCL also improves the top-1 accuracy to $71.4\%$ on the ImageNet over 800 epochs of pretraining.  Although a better result may be obtained by finetuning the hyperparameter
and the data augmentation \cite{tian2020makes}, we stick to the same experimental setting in the previous methods \cite{chen2020simple,chen2020improved} for a direct comparison.

%We make a direct comparison between the KSCL and the MoCo v2  by running both on the same hardware platform with the same set of software such as CUDA10, pytorch v1.3 and torchvision 1.1.0 (used in the data augmentation that plays a key role in the contrastive learning). With 200 epochs of pretraining, the same top-1 accuracy has been achieved on MoCo v2. However, its rerun result over 800 epochs is slightly lower than the reported result (71.1\%) in literature \cite{chen2020improved}, and this may be due to different versions of deep learning frameworks and drivers that could cause variations in the model performance.

%From the results in Table~\ref{tb:imagenet}, the KSCL with $K=5$ and $\rho=40\%$ improves the top-1 accuracy to $68.8\%$ $71.4\%$ on the ImageNet over 200 and 800 epochs, respectively. Although a better result may be obtained by finetuning the hyperparameter
%and the data augmentation \cite{tian2020makes}, we stick to the same experimental setting in the previous methods \cite{chen2020simple,chen2020improved} for a direct comparison.

\begin{figure*}[t]
%\captionsetup[subfigure]{singlelinecheck=false}
    \centering
    \begin{subfigure}[c]{0.95\textwidth}
        \includegraphics[width=\textwidth]{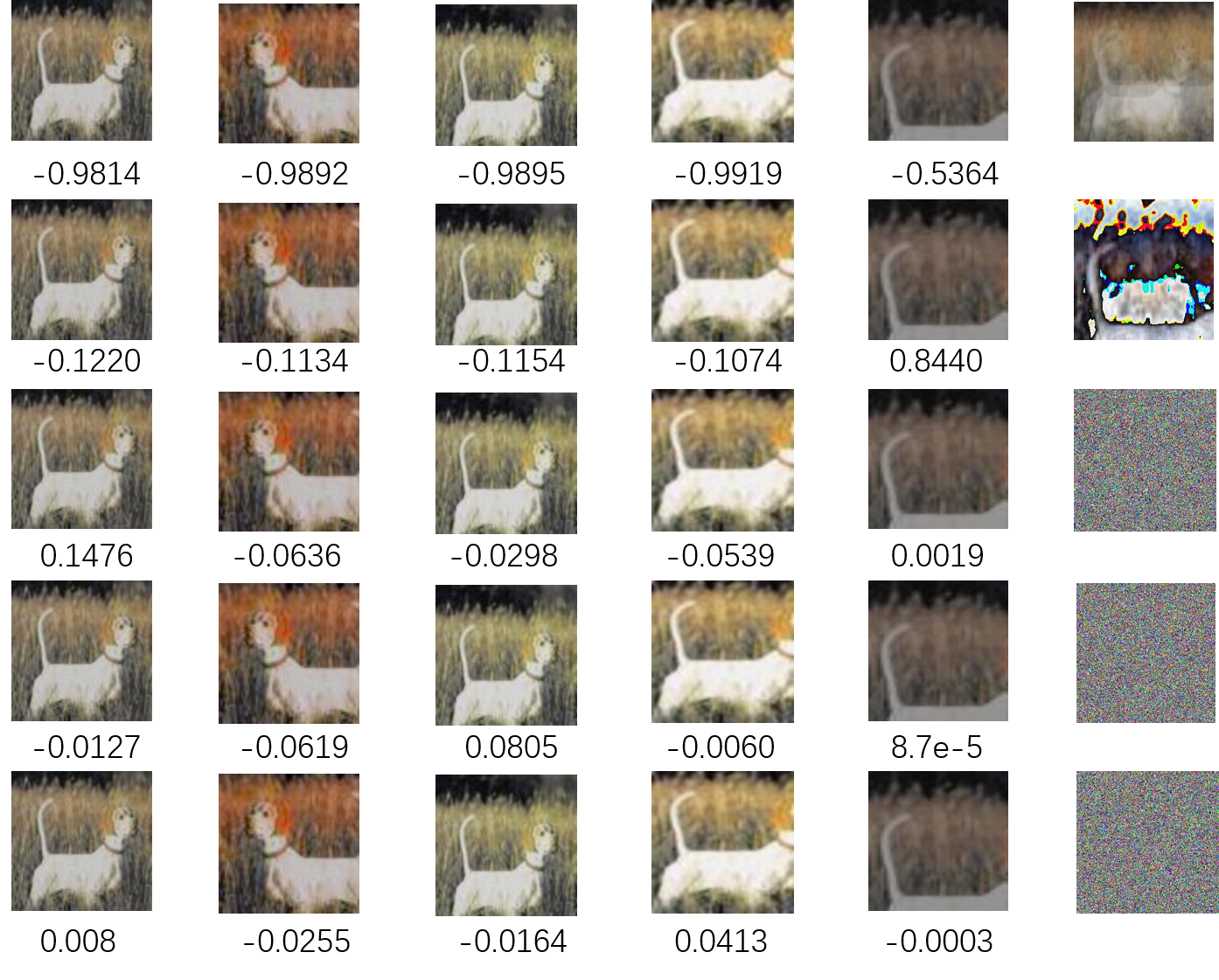}
        %\caption{Auto-Encoding Data (AED)}
    \end{subfigure}
    \caption{The learned basis in an instance subspace. Each of the first five columns is an augmented image from an instance, and the last column is the basis images each of which is synthesized as a linear combination of the five augmented images weighted by the inner product with the corresponding eigenvector in the embedding space.  }\label{fig:eigs_cat}
\end{figure*}

We also visualize the learned basis images in Figure~\ref{fig:eigs_cat}. The last column presents the basis images spanning the underlying instance subspace for a "cat" image. The weight beneath each image is the inner product between the decomposed eigenvector and the embedding of the corresponding augmentation, and each base is a weighted combination of the augmented images in the row. The results show that two bases suffice to capture the major variations among the five image augmentations, while the remaining three only model the minor ones that can be discarded as noises.

\subsection{Impacts of $K$ and $\rho$ on Performance}
We also study the impact of different $K$'s and $\rho$'s on the model performance. Table~\ref{tb:eig} shows the top-1 accuracy under various $K$'s and $\rho$'s. When $K=1$, it reduces to one-shot contrastive learning which is similar to the MoCo v2. The difference $67.2\%$ vs. $67.5\%$ between the KSCL ($K=1$) and the MoCo v2 is probably because we did not fine-tune the temperature $\tau$ for the projection length to optimize the KSCL.

The accuracy increases with a larger number of $K$ augmentations per instance and a smaller value $\rho$ of perceived eigenvalues.  This implies that eliminating the minor noisy variations (as illustrated in Figure~\ref{fig:eigs_cat}) with a smaller $\rho$ could improve the performance. Further growing $K$ only marginally improves the performance. This is probably because the data augmentation adopted in experiments is limited to those used in the compared methods for a direct comparison.  Applying more types of augmentations (e.g., jigsaw and rotations) may inject more intra-instance variations that encourage to use a larger $K$. However, studying the role of more types of augmentations in contrastive learning is beyond the scope of this paper, and we leave it to future research.

%Table~\ref{tb:eig} also shows the computing time used to train the KSCL per epoch. The computing cost increases sublinearly with $K$, due to the feed-forward inference of more augmentations. However, the percentage of preserved eigenvalues does not affect the cost much; indeed the spectral decomposition does not cost too much in the KSCL.

%After unsupervised pretraining with $200$ epochs, the KSCL and the MoCo v2 with a batch size of $256$ are still $2.1\%$ and $0.9\%$ higher than the SimCLR with a batch size of $8,192$ under the same number of $200$ epochs.

%With the same batch size of 256, the KSCL achieves a top-1 accuracy of 68.7\% with $K=5$ augmentations and $\rho=40\%$ of preserved eigenvalues, which is $6.8\%$ and $1.7\%$ higher than SimCLR and MoCo v2 with the same number of epochs and batch size, respectively.

\begin{table*}
\centering
\caption{The top-1 accuracy of the proposed KSCL with varying $K$'s and $\rho$'s under 200 epochs of pretraining on ImageNet. The ResNet-50 backbone was pretrained with two-layer MLP head by applying the same combination of enhanced data augmentations used in SimCLR including stronger color distortion and blurring for a fair comparison. The top-1 accuracy is obtained by training a single-layer linear classifier upon the pretrained features.  We also compare the computing time used to train the KSCL per epoch in eight V100. Note when $K=1$, $\rho$ need not be set as it becomes a trivial case of an one-shot contrastive learning.}\label{tb:eig}\vspace{2mm}
 \begin{tabular}{cc|cc|cc} \toprule
$K$&$\rho$ &epochs & batch size&top-1 accuracy&time/epoch (min.)\\ \midrule
1 & -- & 200 & 256 & 67.2&16\\
3 & 40\%  & 200 & 256 & 68.5&26\\
5 & 40\%  & 200 & 256 & 68.8&37\\
5 & 90\% &  200 & 256 & 68.4&37\\\bottomrule
\end{tabular}
\end{table*}

\subsection{Results on VOC Object Detection}

Finally, we evaluate the unsupervised representations on the VOC object detection task \cite{everingham2010pascal}. The ResNet-50 backbone pretrained on the ImageNet dataset is fine-tuned with a Faster RCNN detector \cite{ren2015faster} end-to-end on the VOC 2007+2012 trainval set, and is evaluated on the VOC 2007 test set. Table~\ref{tb:voc} compares the results with both the MoCo models. Under the same setting, the proposed KSCL outperforms the compared MoCo v1 and MoCo v2 models. The SimCLR model does not report on the VOC object detection task in \cite{chen2020simple}.

\begin{table*}
\centering
\caption{The comparison between the proposed KSCL ($K=5$ and $\rho=40\%$) and the MoCo models. The pretrained ResNet-50 backbone was transferred to train on VOC 2007+2012 trainval set with a Faster RCNN detector end-to-end, and evaluated on the VOC 2007 test set. The COCO metrics were adopted to evaluate the performance.}\label{tb:voc}\vspace{2mm}
 \begin{tabular}{c|cc|ccc} \toprule
Model&epochs&batch size&AP$_{50}$&AP&AP$_{75}$\\ \midrule
MoCo v1 & 200 & 256 & 81.5 & 55.9 & 62.6 \\
MoCo v2 & 200 & 256 & 82.4 & 57.0 & 63.6 \\
MoCo v2 & 800 & 256 & 82.5 & 57.4 & 64.0 \\\midrule
KSCL & 200 & 256 & 82.4 & 57.1 & 63.9 \\
KSCL & 800 & 256 & {\bf 82.7} & {\bf 57.5} & {\bf 64.2} \\\bottomrule
\end{tabular}
\end{table*}

\section{Conclusion}\label{sec:concl}
In this paper, we present a novel $K$-shot contrastive learning to learn unsupervised visual features. It randomly draws $K$-shot augmentations and applies them separately to each instance.  This results in the instance subspace modeling  how the significant factors of variances learned from the augmented samples can be linearly combined to form the variants of an associated instance. Given a query, the most relevant samples are then retrieved by projecting the query onto individual instance subspaces, and the query is assigned to the instance subspace with the shortest projection distance.  The proposed $K$-shot contrastive learning combines the advantages of both the inter-instance discrimination and the intra-instance variations to discriminate the distinctions between different instances.  The experiment results demonstrate its superior performances to the state-of-the-art contrastive learning methods based on the same experimental setting.

\ifCLASSOPTIONcaptionsoff
  \newpage
\fi



%\bibliographystyle{IEEEtran}

\bibliographystyle{IEEEtran}
\bibliography{KSCL-MIA}

\end{document}